# Patient Similarity Analysis with Longitudinal Health Data

Ahmed Allam[1,2,+], Matthias Dittberner[1,2,+], Anna Sintsova[1,2], Dominique Brodbeck[3], Michael Krauthammer[1,2,4,*]

[1] Department of Quantitative Biomedicine, University of Zurich, Zurich, Switzerland
[2] Biomedical Informatics, University Hospital of Zurich, Zurich, Switzerland
[3] Institute for Medical Engineering and Medical Informatics, University of Applied Sciences and Arts Northwestern Switzerland
[4] Yale Center for Medical Informatics, Yale University School of Medicine, New Haven, Connecticut, USA
[+] Joint-first authors
[*] corresponding author: michael.krauthammer@uzh.ch

**Abstract**

Healthcare professionals have long envisioned using the enormous processing powers of computers to discover new facts and medical knowledge locked inside electronic health records. These vast medical archives contain time-resolved information about medical visits, tests and procedures, as well as outcomes, which together form individual patient journeys. By assessing the similarities among these journeys, it is possible to uncover clusters of common disease trajectories with shared health outcomes. The assignment of patient journeys to specific clusters may in turn serve as the basis for personalized outcome prediction and treatment selection. This procedure is a non-trivial computational problem, as it requires the comparison of patient data with multi-dimensional and multi-modal features that are captured at different times and resolutions. In this review, we provide a comprehensive overview of the tools and methods that are used in patient similarity analysis with longitudinal data and discuss its potential for improving clinical decision making.

# 1  Introduction

Driven by the expansion of health services and the broad introduction of the Electronic Health Records (EHRs) in the last decades, there has been a massive increase in the amount of available digital medical data. The volume of data was estimated to be 153 Exabytes ($10^{18}$ bytes) in 2013 and projected to reach 2,314 Exabytes in 2020[1], or in other words, roughly 300 Gigabytes per person on average. These data, which are currently underutilized, are a promising and abundant resource for uncovering new medical knowledge that could potentially improve patient health outcomes.

For the last three decades, medical professionals have turned to randomized controlled trials (RCTs) as the most reliable source of clinical evidence. While RCTs remain the cornerstone of medical decision making, there is increasing consensus that the vast EHR data repositories and the medical knowledge generated from them should also be regarded as a complementary source of clinical evidence[2,3]. These efforts are part of what is now called the learning health care system[4], which searches for clinical evidence in real world data by identifying novel correlations or adjusting known ones. Newly discovered clinical signals from data-driven analysis can then be validated using RCT standards.

While RCTs usually look at 2 or 3 timepoints, the increased availability of longitudinal health data is making it possible to learn from a complete patient journey (i.e. a chronological record of medical visits, symptoms, tests, procedures and outcomes). Using these data, patient similarity analysis aims to classify patients into medically relevant clusters in order to gain insight into underlying disease mechanisms and facilitate precision medicine. Specifically, this clustering of patient journeys can identify common (as well as previously unrecognized) disease trajectories that lead to a specific outcome. Moreover, in precision medicine, patient similarity analysis can be used to improve patient outcome prediction. This is based on the idea that outcome prediction models trained on data from patients similar to the index patient are more accurate than those trained on all available data. Both of these applications of patient similarity analysis are likely to contribute to clinical decision making in the future.

Medical data are often innately noisy and irregular with high rates of missing data, which makes comparison of patient journeys a difficult undertaking. Here we present a systematic review of papers that use patient similarity analysis to derive novel clinical insights from historically collected longitudinal health data such as EHRs or high-resolution Intensive Care Unit (ICU) data. We highlight the methods used in patient similarity analysis with regards to (1) Data abstraction methods, (2) Similarity metrics used, and (3) Algorithms applied across different application domains. Our investigations go beyond existing reviews that are focused on the overall opportunities, challenges, and the use of machine learning (i.e. specifically deep learning) in medicine (beyond the scope of current work)[5–10], or specific aspects of either longitudinal data analysis or patient similarity analysis[11–13].

# 2  Methods

We conducted a systematic literature review of prior work discussing patient similarity analysis of patient journeys. We used combination of relevant keywords to search IEEE Xplore, ACM Digital Library, arXiv, and PubMed for studies that were focused on the analysis of patients' longitudinal data (i.e. patient journeys) and included the notion of "similarity/dissimilarity", "matching", and similar terms (list of exact keywords used is found in Supplementary S1). Figure 1 depicts the literature selection workflow. Our search strategy identified 125 papers discussing patient similarity assessment in longitudinal health data between the years 1995 and 2019. While there was an increase in papers over the years, studies on personalized predictions showed the biggest growth (Figure 2). In Supplementary S2, we provide a breakdown of each paper according to specific application area, used methods and dataset sources, as well as the chosen outcome and evaluation metric.

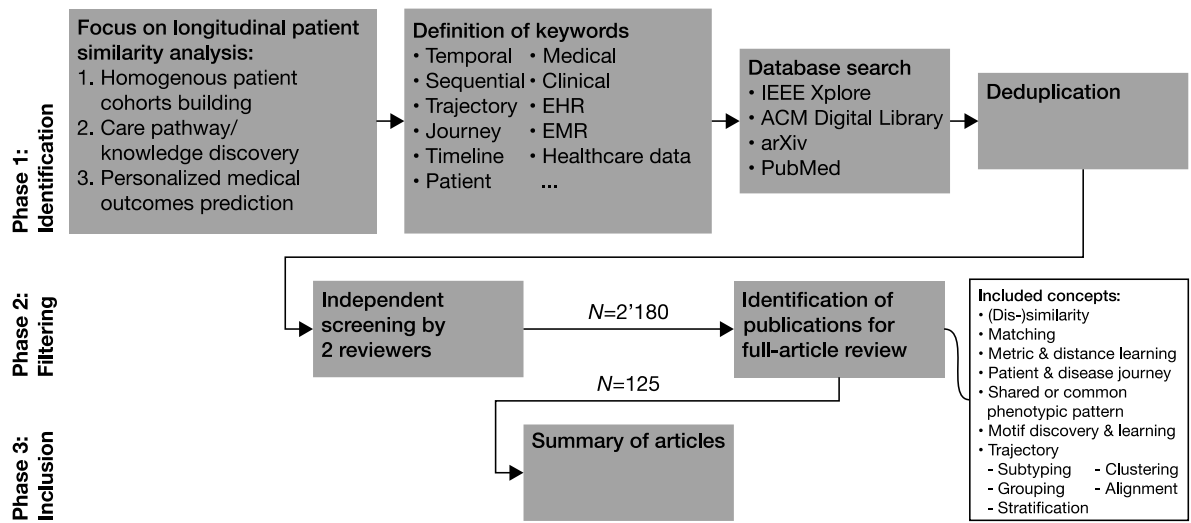

Figure 1. Literature selection workflow

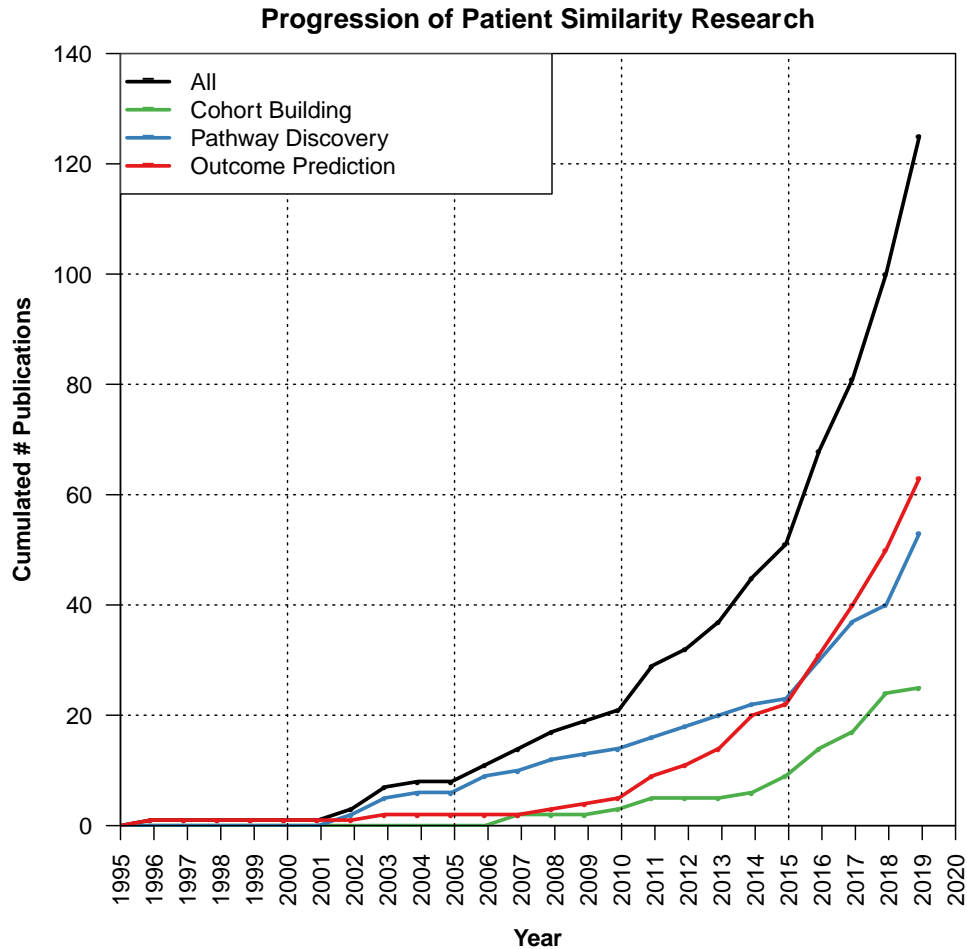

Figure 2. Progression of research on patient similarity assessment with longitudinal health data

## 3. Results

*Basic workflow:* The reviewed papers reveal a prototypical workflow for patient similarity assessment (Figure 3). At forefront is the need to perform **data abstraction** (see section 3.1) for reducing the complexity of patient journeys (i.e. abstracting the timeline into vectorial representations). Using this abstraction, we can either explicitly define a **similarity metric** (see section 3.2) for the grouping of journeys (using distance measures such as Euclidean distance and/or dynamic time warping, sequence alignment and distance metric learning), or apply **algorithms** (see section 3.3) that implicitly incorporate similarity (i.e. kernel methods, gaussian processes) for use in various application domains.

*Application domains*: Moreover, we can categorize the identified application domains into three high-level categories - **cohort building**, **pathway discovery**, and **personalized outcome prediction** (Figure 3, bottom). Assembling patient journeys into personalized cohorts based on similarity to an index patient minimizes the bias and is the basis for both the grouping of journeys into common pathways (for the discovery of latent disease state, endotypes or typical care patterns) and personalized outcome prediction (for learning about treatment options that worked for a majority of patients in a cohort).

*Data Sources:* Commonly used datasets reported in the reviewed papers were based on longitudinal EHR data as in[14–19], ICU data[20–28], claims/administrative[20,29] and registry data[30]. Moreover, multiple studies used freely accessible datasets such as MIMIC II and III[31], Alzheimer's Disease Neuroimaging Initiative (ADNI)[32], eICU-CRD (critical care database)[33], challenge datasets such as Parkinson's Progression Makers Initiative (PPMI) challenge[34], PhysioNet computing cardiology challenge[35], TREC Medical Track collection data[36], ECML/PKDD discovery challenge[37], UCI benchmark data[38] and other miscellaneous datasets[39–41].

*Evaluation Metrics:* Papers that focused on outcome-related tasks predominantly presented results using the Area Under the Receiver Operating Characteristic curve (AUROC), F-measure, precision, recall, accuracy, root mean squared error and Matthews Correlation Coefficient. Papers focusing on similar patients' retrieval included precision/recall/specificity@k metric, normalized discounted cumulative gain (NDCG). Papers focusing on clustering approaches used Purity, Mutual Information, Rand Index, Silhouette analysis, significance testing and distribution analysis of dominant features in identified clusters. Qualitative analysis was also used. Here authors and/or physicians commented on the obtained results and qualitatively contrasted the findings to what is known in the literature.

# Patient Similarity Analysis

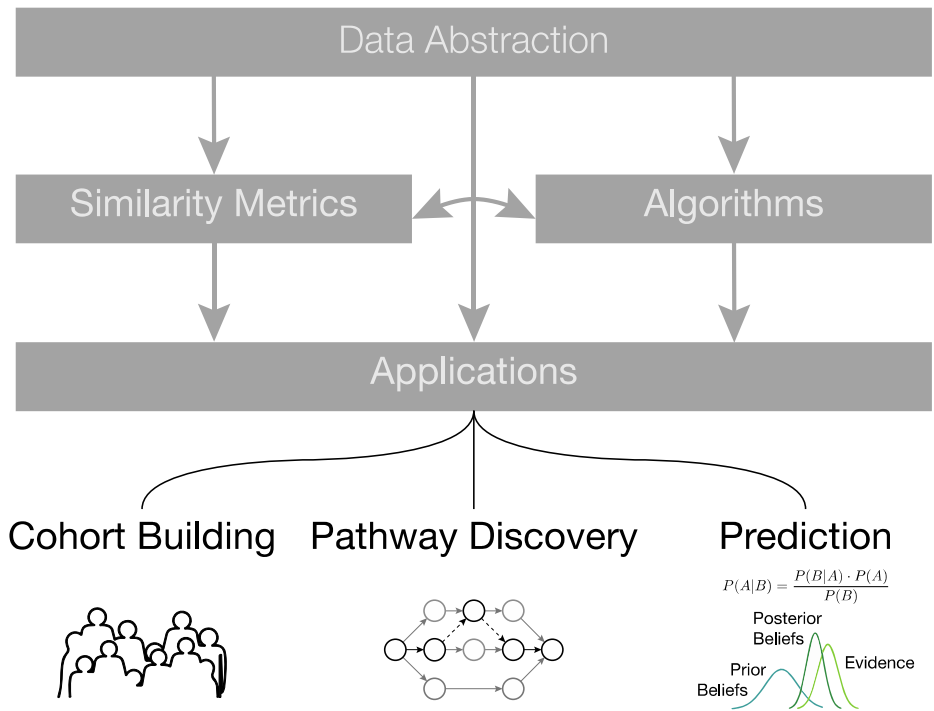

Figure 3. Basic processing schema (top, grey boxes) and application domains (bottom) for patient similarity analysis with longitudinal health data.

## 3.1 Data Abstraction

Patient data is described by high-dimensional features, where each feature is captured at different times and resolutions, and with different levels of completeness. This results in noisy and irregular data with high rate of missing values. Hence, there is no one-to-one correspondence between patient features, making straightforward assessment of similarity between patients a difficult task. Consequently, the use of **data abstraction**, in which the patient data are transformed into a possibly lower-dimensional and non-sparse representation, is a promising way for proper journey comparison. Below we describe data abstraction methods that have been successfully applied to either EHR or high-throughput ICU streaming data. While many temporal abstraction methods have been applied, recent research has focused on using neural networks to automatically learn the best data representation.

**Symbolic Temporal Abstraction**
Symbolic temporal abstraction is the process of transforming the raw time-based measurements into interval-based abstraction focused on symbolic representation. This can be a powerful way to represent irregular EHR data. These interval-based representations are known as time interval related patterns (TIRPs). In the context of patient similarity analysis, this transformation can be done using (1) pre-specified thresholds on the measured variable or binning approaches, or (2) statistical algorithms such as symbolic aggregate approximation (SAX)[42].

*Thresholding*: As one example, Moskovitch et al.[42] constructed TIRPs by mapping the raw time-stamped medical data into conceptual representations ("HbA1c increasing", "stable diabetes medication dose", *etc*.). The authors enumerated and considered TIRPs that have frequency above a given support threshold using modified version of Allen's relations (i.e. before, meets, overlaps, finish-by, contain, start-by, equal and their inverse). The algorithm was tested using medical records from diabetic patients collected over a five years timespan. They discovered frequent TIRPs that identified clusters of patients with similar temporal relations among different variables. Further work[41] showed that increasing the minimal support (i.e. minimum threshold) led to higher consistency of the discovered TIRPs, as did the use of a Semantic Adjacency Criterion (SAC) as a formal defined constraint. Moskovitch and colleagues[43,44 45] then also applied this technique for outcome prediction. Using eight cohorts of patients derived from the New York-Presbyterian/Columbia University Medical Center Clinical Data Warehouse, the authors used their framework to predict eight commonly performed clinical procedures targeting different organ systems. The authors showed that training a model using the detected TIRPS outperformed non-temporal models incorporating symbolic representation or TIRPS with Boolean representations of patients' timelines.

Bonomi et al.[28] built on the TIRPs concept, proposing a similarity measure that takes into account noise in the TIRP representation at both the value and temporal relations level. Applying their work on MIMIC-III dataset, the authors constructed time interval sequences consisting of diagnosis codes, procedures and prescriptions associated to each patient's admission. Given two TIRPS the similarity between them was measured using Hamming distance for both the values and the relations. The authors give examples of real-world TIRPs and their distances, however further work is needed to evaluate the performance of this methodology.

*Symbolic Aggregate Approximation (SAX)*: SAX converts the numerical form of time series into a sequence of discrete symbols (SAX words) according to pre-specified mapping rules. Lin and Li[46] used SAX to convert medial timeseries from an electrocardiogram (ECG) dataset[47] into SAX words and used them to construct a word-sequence matrix. The authors showed that the use of Euclidean distance on this data representation resulted in improved clustering (using hierarchical clustering and k-means) and classification (1-Nearest Neighbor) of ECG time series, compared to clustering based on Euclidean distance or Dynamic Time Warping (DTW) of raw time series data. Zhao et al.[48], explored the use of Piecewise Aggregate Approximation (PAA) together with SAX for detection of adverse drug events (ADEs) in patient EHRs. PAA segments a time series into fixed non-overlapping intervals, and each of these is then represented by the average of all data points time-stamped within the interval. The authors (1) computed PAA of the signal (2) mapped the signal to a discrete symbol using SAX algorithm and (3) defined a shapelet (motif representing time series subsequences) based on the sequence of symbols generated by SAX algorithm. This approach along with Random Forest model showed promising results for detection of ADEs. Ghosh et al.[49] used SAX and sequential contrast algorithm on mean arterial pressure time series data to identify subsequences predictive of acute hypotensive episodes. Morid and colleagues[50] implemented temporal abstraction based on PAA and adjacency-based imputation of missing values and validate their methods in the context of early ICU mortality prediction.

**Data Representation Learning using Neural Networks**

Different approaches have been proposed to automatically learn the best data representation using neural networks. A number of groups implemented autoencoders (AE) to create a fixed-length vector representation of medical data[51]. For example, the AE implemented by Dhamala and colleagues[24] used an encoder-decoder that is based on a gated recurrent unit (GRU) neural network. The proposed framework achieved an improved accuracy for predicting acute hypotensive events compared to state-of-the-art approaches. Zhang et al.[52] proposed a time-sensitive gated RNN (GRU) based AE to model EHR time series data with irregularly sampled time points. Beyond modeling time-irregularity, the authors proposed to jointly model static data, such as demographics or family history, which is typically recorded only once in an EHR record. Using k-means clustering on the representations learned from a longitudinal Parkinson dataset, they demonstrated that their approach generates patient clusters with highly distinct clinical features. To ensure they are accounting for noisy and missing values in the data representation, Bianchi and colleagues[53,54] proposed a deep AE[55] bidirectional recurrent neural network architecture such that pairwise similarity in the input space is preserved using a kernel function. Using an ECG dataset, the authors compared their method to other dimensionality reduction techniques and showed the role of kernel alignment in improving the learned representation in the case of missing data. Beaulieu-Jones et al.[56] presented a semi-supervised learning approach for clustering and progression prediction of Amyotrophic lateral sclerosis (ALS) patients using denoising AEs to construct a representation from aggregated measurements of patients' medical histories. Beaulieu-Jones[26] used autoencoders and long short-term memory RNN (LSTM) as temporal abstraction of patient timelines for unsupervised and supervised modeling of patient care events. Their use of t-SNE [57] with clustering showed meaningful embeddings with patient trajectories linked to 1-year mortality grouping together. Baytas et al.[58] developed a time-aware LSTM (T-LSTM) model using an AE to learn a representation for patients' longitudinal records that were later clustered into clinical subtypes using k-means. Their approach generated meaningful patient sub-types using Parkinson's Progression Markers Initiative dataset.

**Other Data Abstractions Approaches**

Henriques and colleagues[59] proposed a pattern recognition strategy to compute the similarity between two time series using data from physiological signal measurements. They used Haar wavelet decomposition, in which signals are represented as linear combination of a set of orthogonal basis. As a further dimensionality reduction approach, Karhunen-Loève transform was applied to the eigenvectors of the covariance matrix of the wavelet basis. The authors concluded that the resultant compact data representation is particularly suitable for maintenance of historical data records. In another example, Montani et al.[60]. presented a retrieval tool for time series data of dialysis patients, which uses Discrete Fourier Transform for dimensionality reduction and Euclidean distance for similarity assessment.

Sacchi et al. [61] studied longitudinal blood glucose data to identify patients with varying risk profiles. The authors used temporal abstraction to represent time series data into sequence of intervals focusing on three patterns: increasing, decreasing and stationary patterns. Then the transformed sequences were clustered using temporal abstraction clustering to generate patients' risk profiles for experiencing hyper- or hypo-glycemic episodes during ICU stays. Batal et al.[62] used time interval-based representation similar to Höppner's representation for identifying patterns associated with the detection of adverse medical events in diabetic and post-cardiac surgery patients. Other reported data abstraction approaches were based on wavelet transform[63], description logic[64], knowledge-based temporal abstraction[65], residual deviance representation[66], time-annotated sequences[67], shapelets[68], fuzzy space-time windowing[69], and temporal profile graphs[70–72].

## 3.2 Similarity Metric

Data abstraction does not solve another fundamental challenge of patient similarity analysis, which is the choice of the metric that would allow for a meaningful understanding of whether two patients are comparable with respect to their care pathway and disease outcome. A large body of work addresses the different aspects of selecting an appropriate similarity metric, with more recent publications, as observed in the Data Abstraction section above, focusing on Neural Networks for similarity metric learning.

**Heuristic and Distance-based Similarity Metrics**
The most straightforward approach to designing a similarity metric is to compare known clinically relevant patient features. For example, Chan and colleagues[73] presented a "similarity-based" machine learning approach for outcome prediction in hepatocellular carcinoma (HCC). The authors applied multiple custom similarity measures to the 14 known risk factors of HCC. Based on these measures, an SVM classifier showed best performance in predicting if each pair of patients is similar or dissimilar based on their survival time. Gottlieb et al.[74] built personalized cohorts based on 10 similarity measures (such as co-occurrence frequency and nearest common ancestor in the international classification of diseases (ICD) coding hierarchy, ECG and blood tests similarities) to predict ICD codes upon hospital discharge. Wang et al.[75] also used heuristic similarity measures (such as diagnosis, vital signs, and lab test similarities) together with an agglomerative nesting algorithm for patient clustering and cohort identification. Sun and colleagues[76] proposed to divide a medical time series into discrete treatment episodes, represented as tables of physician orders consisting of one or more drugs and dosage details. They then constructed a heuristic similarity metric that calculated the weighted average similarity between treatment episodes of two time series using sequential rules, such as comparing drug names and drug delivery routes. Using a Density Peaks based clustering approach on EHR data, they demonstrated the ability to find prototypical treatment regimens.

Different distance metrics have been used to measure patient similarity. As mentioned in the Data Abstraction section, many studies use Euclidian distance following data transformation. Alternatively, cosine similarity is a popular choice of a distance metric[77–79]. Kim and O'Reilly[80] proposed a scalable fast retrieval system for patients with similar physiological waveforms based on locality-sensitive hashing (LSH) method[27]. The similarity between sequences was assessed using $l_1$ distance and their method was tested on arterial blood pressure time series dataset extracted from the MIMIC-II database. Furthermore, Hielscher and colleagues proposed the use of Heterogeneous Euclidean Overlap Metric (HEOM), which is able to deal with missing values in times series data, for classifying patients with hepatic steatosis, with the similarity measure being used for both feature selection and classification[81,82].

**Similarity based on Dynamic Programming**

*Smith-Waterman algorithm*

A number of studies used a variation of Smith-Waterman local alignment algorithm as the basis for their similarity metric. For example, Sha et al.[83] constructed sequences from patients' laboratory tests and computed local alignment between pairwise sequences to find similar patient journeys. The authors used their approach with k-nearest neighbor (k-NN) for mortality prediction of ICU patients with acute kidney injury and sepsis. Syed and Das[84] also used Smith-Waterman for local alignment, as well as Needleman-Wunsch for global alignment, to find similar trajectories in medical time series data. Their adaptation of alignment algorithms accounted for the time between events while penalizing missing events. The authors used their algorithm to automatically identify the chemotherapy protocol of breast cancer patients by aligning the patient journey to a known treatment protocol. Hajihashemi et al.[85,86] reported another variation on the Smith-Waterman local alignment algorithm for improved prediction of change in patient's health status based on temporal sensor data collected from aging facility. Authors computed a pairwise similarity between a new recording and all "normal" sensor recordings in a training set. The computed similarity is then compared to fitted Gamma distribution of pairwise similarities of all normal sensor recordings to determine if a recording was normal or abnormal.

*Dynamic Time Warping*

Dynamic Time Warping (DTW) is also frequently implemented to assess similarity between time series sequences. For example, Zhang et al.[87,88] implemented DTW for matching temporally ordered sequences of patient visits, each visit being represented by MetaMap-encoded[89] and vectorized clinical notes. They then used multiple clustering algorithms to group sequences of patient visits. In all of these approaches, the authors used a common sequence reflecting many patient's features together. In contrast, Tsevas and Iakovidis[90] used DTW in a multi-modal setting where they constructed separate temporal sequences for each patient feature. Similarity matrices computed for individual sequences were then merged using a fusion scheme. The authors applied their method to a hepatitis dataset[37] to successfully retrieve relevant cases based on a query case. Salgado et al.[91] described a fuzzy clustering algorithm (based on fuzzy c-means) that uses DTW on time-variant (physiological variables), as well as time-invariant (demographics) data. The authors used their approach for predicting vasopressors administration in patients with pancreatitis and risk of mortality for patients with septic shock. Goyal and colleagues[92] presented a matching approach based on subsequence alignment to compute similarity among patients using DTW. The authors contrasted their approach to global-based alignment methods and tested their model on the task of stratifying patients with risk of developing Alzheimer's disease. The results support the use of DTW over global alignment methods. Interestingly, Hoogendoorn et al.[93] compared a similarity-based approach (prediction model that used similar patients for outcome prediction) to a "global" predictive model using the whole cohort to predict ICU mortality. The similarity-based model used DTW with Euclidean distance as a similarity measure and k-NN for predicting mortality risk. Testing their work on MIMIC-II dataset, the authors found this similarity-based model achieved lower performance compared to a global approach.

**Interactive Similarity Learning**

Some papers put the focus on learning an appropriate similarity metric using expert input. In this case, the similarity labels are provided by the physicians, and the similarity metric is learned algorithmically.

Sun and colleagues[94] presented a system comprising of a suite of methods for performing supervised similarity metric learning. The system incrementally updates an existing learned metric using physicians' feedback and used Comdi[95] for aggregating multiple learned similarity metrics from several physicians. The authors demonstrated that their system retrieves similar patients with higher precision than baseline systems. Given feature vectors representing patient's measurements and similarity labels provided by expert physicians, Sun et al.[20,21,96] learned a generalized Mahalanobis distance, where the precision matrix captures the correlations between different features, such that patients with different labels are far from each other, and patients with identical labels are close together. A model using localized metric learning with features represented in the wavelet domain (i.e. using wavelet coefficients) achieved the best performance in both classification and retrieval tasks. Zhan and colleagues[97] also discussed the use of a generalized Mahalanobis metric as similarity learning approach in conjunction with Group LASSO for feature selection. The algorithm was tested on the UCI benchmark dataset as well on a longitudinal real-world clinical dataset to successfully retrieve similar patients. Wang and colleagues[98] presented a system in which physicians' feedback is partially used as labels in a semi-supervised learning process. Similar to[95], the procedure projects a patient feature vector to lower dimensional space in which pairwise similarity between patients is computed. A prediction using the top retrieved similar patients outperformed the baseline methods in predicting congestive heart failure 6 month later. Ng and colleagues[99] also proposed the use of locally supervised metric learning for retrieving sets of patients similar to a given index patient. They then trained a logistic regression using the top similar patients from two classes (i.e. positives and controls) to predict the probability of developing diabetes. Additionally, Wivel et. al.[40] discussed a more guided and iterative approach, which relied on expert discussions about the appropriate clinical, functional and biological variables that are needed for assessing patient similarity. This strategy was used to establish and compare individual patient trajectories using data from two randomized controlled trials (RCTs) assessing the efficacy of belimumab in patients with systemic lupus erythematosus (SEM). The authors reanalyzed the RCT data constructing variable-specific patient journeys, categorizing each journey into "improved", "worsened" or "stable". The paper highlights the difference between traditional single time-point reporting (typical for RCTs) and information learned from patient journey analysis, which reveals refined subgroups within the responder and non-responder categories.

**Using Neural Networks to Learn Similarity Metric**

Many papers proposed the use of neural models for computing similarities among patient journeys. Suo et al.[15] investigated a time fusion convolutional neural network (CNN) framework to simultaneously learn patient representation and pairwise similarity between two patients' time series. Building on the work of Zhu et al.[100], they proposed a similarity computation procedure using a multilayer perceptron (MLP) and segmented one-sided convolution neural network. Once the model was trained, multiple prediction models were trained using cross-entropy loss and in a later paper[101] using a triplet loss function on personalized cohorts based on the learned similarity representation. The models' performance was compared to a "global" logistic regression model. Using claims data, they showed that personalized prediction bests prediction based on the global model, and that a CNN-derived similarity representation outperforms a baseline representation. Che et al.[102] built a similarity-based model based on GRU RNN with a modified version of DTW using a 2D-RNN architecture with ranking loss function to Parkinson disease patient sub-groups. Their proposed personalized approach also outperformed "global" predictive models and showed to be robust to noisy and irregularly sampled data. Finally, Zhu et al.[100] adapted ideas from word embedding, deriving fixed-length vector representation from EHRs by medical concept embedding (word2vec using skip-gram model). For similarity metric learning, the authors implemented a CNN, where wide feature maps and pooling were applied to get a fixed-length vector representation for each patient. These representations were then used to calculate pairwise similarity metric between patients. Then the fixed vector representations were joined along with the computed similarity into one vector that was passed to *softmax* layer for predicting if two patients were similar. Using longitudinal EHR data, the authors showed that this deep learning approach achieved significantly better results compared to unsupervised approach using R-Vector coefficient and distance covariance coefficient[103,104].

### 3.3 Algorithms

In addition to the work done on data abstraction and similarity metric learning, a significant body of work has been dedicated to developing methods for modeling and processing of patient journeys to learn associations, identify motifs and patterns that are discriminative for predicting or clustering those journeys. In this section, we discuss the fundamental algorithm categories that are used in cohort building, patient stratification, and patient outcome prediction from longitudinal health data.

**Sub-pattern discovery algorithms**

These algorithms focus on the effective clustering and pattern detection from patient journeys. A good example of a sub-pattern discovery algorithm is multiscale structure matching (MSM), which compares two sequences by partially changing observation scales to examine similarity between them from long- and short-term perspectives. To calculate similarity, the approach attempts to find the best pair of sub-segments, even across scales, that minimizes total difference between the sequences. Hirano et al.[105,106,115,116,107–114] in a series of papers, discussed the use of MSM-based algorithms for analyzing time-series of laboratory values. Using data form hepatitis B and C patients, they presented a qualitative evaluation of the results, identifying times-series indicative of liver damage, as well as meaningful clustering of patients based on discovered patterns.

Other groups implemented nonnegative matrix factorization (NMF) approaches for pattern discovery in EHRs[117–119]. For example, Wang and colleagues[118,119] presented their system OSC-NMF (One-Sided Convolutional Nonnegative Matrix Factorization) for pattern discovery in longitudinal medical records. The method is devised to deal with data heterogeneity (multidimensional/multimodal data), sparsity as well as irregularity. The algorithm computes a one-sided convolution and minimizes $\beta$-divergence between the original event matrix representation and the discovered set of one-sided convolutional patterns. The authors tested their algorithm using synthetic and real data, discovering meaningful and predictive temporal patterns in a diabetes dataset.

A number of clustering methods have been applied to the problem of patient journey classification, including hierarchical clustering, k-means[120], constraint-based clustering algorithms[121], and spectral clustering[17].

Other groups tried to differentiate between journeys of sick and normal patients using ICU and/or vital sign data using the Hurst exponent and fractal to quantify the ruggedness of the data[122], and estimating the distribution of normal and abnormal journeys using kernel density estimation[123].

Other algorithms applied to pattern discovery problem using longitudinal patient data included association rule mining algorithm[124], functional dependencies mining [125], tensor factorization algorithm[126], and regression-based trend templates algorithm[127].

**Hidden Markov Models**

HMMs are popular choice for patient journey modelling and was used for both clinical pathway discovery and outcome prediction. For example, Galagali et al.[128] presented a continuous-time HMM (CT-HMM) that models patient journeys to infer the progression of a hidden disease state using finite discrete state space. Under this CT-HMM model, a likelihood of a given journey can be computed, offering a measure to assess similarity among patients, with more probable journeys being more similar. The authors used mixture model to cluster patients into disease subtypes by maximizing the joint probability of subtype assignment and the conditional probability of patient journeys. They demonstrated the utility of their model by discovering subtypes of progression to hemodynamic instability (requiring cardiovascular intervention) in a patient cohort from a multi-institution ICU dataset.

A number of groups implemented Bayesian Hidden Markov Model (B-HMM)[17,129–131]. The paper by Huang et al.[129] segmented each patient's journey into set of epochs, i.e. sets of treatment events. The goal was the identification of latent treatment topics for each epoch and the transition of topics in a given patient's journey. The authors constructed a B-HMM to represent patients' journeys with Dirichlet priors placed on the transition parameters and emissions distributions. A likelihood of patient journeys is used to calculate similarity among patients, and agglomerative hierarchical clustering algorithm was used to cluster patients into homogeneous groups. Using data from Chinese EMR for patients spanning 9 years following unstable angina treatment, the authors found that their method achieved a better performance compared to baseline models.

Zhang and Padman[132] first clustered patients into homogenous groups using longest common subsequence as distance metric. Then for every group a first order HMM was trained to infer most probable clinical pathways given sequences of multiple laboratory tests observations and other patients characteristics. Using EHR data from chronic kidney disease patients, the authors identified most probable clinical trajectories.

In a series of papers, Lehman et al.[25,133,134] investigated the use of autoregressive HMM, or switching vector autoregressive process (SVAR) to learn shared dynamics from time series measurements. The goal was to identify recurring time series segments within patient's measurements that may be shared across the whole cohort. The authors showed that modeling blood pressure measurements using SVAR outperforms a model using Simplified Acute Physiology Score (SAPS-I) scores for mortality prediction within 24 hours of a ICU stay, and a SVAR-based model of heart rate and blood pressure improves sepsis detection when combined with SAPS-I scores[25,133,134].

**Other Graphical Models**

Besides HMMs, a number of other probabilistic graphical models were used. As with HMMs, these were applied to both knowledge discovery and outcome prediction domains. The work of Schulam et al.[16] attempted to identify subgroups of patient journeys, starting with the assumption that clinical severity markers can reflect disease subtypes. The authors proposed a comprehensive hierarchical model that accounts for variability stemming from factors that are not linked to the disease mechanism (i.e. covariate-dependent as well as individual-specific variability). The authors evaluated their model on a scleroderma data set by predicting marker measurements at unobserved time points and performing a qualitative subtype analysis. The model had better prediction performance than selected baseline models and identified novel disease subtypes for different markers. Schulam and Arora[30] proposed a probabilistic model (disease trajectory map, DTM) to map clinical marker trajectories into a lower-dimensional vector space to capture latent similarity between patients using Euclidean distance. The DTM model was based on reduced-rank formulation of linear mixed models[135] where the inference was done using stochastic variational inference[136]. Low-dimensional representations were consistent with medical literature and showed an association with important clinical outcomes. Zeldow et al.[137] discussed a Bayesian generalized linear model (GLM) with an enriched Dirichlet process (EDP) for regressing lab tests on baseline covariates such that the model predicts lab measurements at any desired time points in a time series. The authors tested their approach by predicting lab measures indicating diabetes status exactly one year following initiation of a second-generation antipsychotic (SGA) drug. As model optimization implicitly involved clustering of patient-specific trajectories of lab measures, they also demonstrated the ability to find meaningful subgroups of lab measurement trajectories after SGA administration. Alaa et al.[138–140], in a series of papers, presented a methodology for risk prognosis for critical care patients. Their approach first discovers latent patient subtypes in unsupervised way using expectation maximization (EM) algorithm trained on EHR data. The authors then trained a Gaussian Process (GP) for every discovered subtype using their clinical time series measurements to capture the physiology of every sub-type. The algorithm outperformed state-of-the-art scores in predicting ICU admission. Chung and colleagues[23] investigated a mixed-effect based model for outcome prediction. Their approach aimed at capturing a global trend across all patients through the use of a multi-layered RNN network and a personalized patient-specific trend using GP. The model showed superior performance for disease classification and length of stay prediction compared to baselines using RNN or GPs alone.

Marlin et al.[141] discussed clustering of pediatric ICU time-series data by discretizing time into hour-long intervals using PAA. A Gaussian mixture model was used to cluster the timeseries with an informative prior distribution (i.e. square exponential kernel based gaussian prior) on the mean parameters to alleviate the sparseness of measurements in the dataset. The posterior distribution of the latent variable representing the cluster membership allowed for assigning each patient to the most likely cluster. Moreover, the authors used Bernoulli cluster model to predict mortality per cluster basis using posterior weighted data. Suresh et al.[142] presented a two-step framework in which they learned patient subtypes and then predicted the outcome for each patient subgroups. They used an LSTM-based RNN sequence-to-sequence (seq2seq) autoencoder for time series representation and then used a Gaussian Mixture model to identify patient subgroups. The authors then used multi-task LSTM model to predict in-hospital mortality using MIMIC-III database.

Other graphical/neural network models used for patient journey modelling included Restricted Boltzmann Machine (RBM)[143], Conditional Restricted Boltzmann Machine[144], Bayesian network model[145] and Hierarchical Dirichlet Process model[146].

**Heuristic Algorithms to Identify Common Disease Trajectories**
The heuristic algorithms described in this section mostly focus on discovering common disease trajectories between patients and attempt to predict the intervention that would change these trajectories. For example, the algorithm proposed by Jensen and colleagues[18] starts by identifying significant correlations between pairs of diagnoses by calculating risk ratio in relation to comparison patients. The correlated diagnosis pairs are extended to full patient trajectories by concatenating pairs sharing the same diagnosis codes. Trajectories are retained and ordered by the number of patients that follow the full trajectory from beginning to end. In a final step, trajectories are clustered using Jaccard Index as a similarity measure. Using EHR data spanning more than 14 years across 6 million Danish patients, their work revealed prototypical pathways leading to septicemia starting from 3 original states (alcohol abuse, diabetes and anemia) and established the trajectory-specific probability of sepsis mortality[147,148]. Starting with patients with no diabetes diagnosis at baseline, Oh et al.[149] constructed typical patient trajectories of developing three comorbidities (hyperlipidemia, hypertension, and impaired fasting glucose) and evaluated the probabilities of such trajectories (and their permutation) using EHR data. The authors identified the most common trajectory of comorbidity development, as well as atypical ones that contributed to increased log odds of developing type 2 diabetes mellitus (T2DM). Al-Mardini et al.[150] discussed a mechanism for proposing patient-specific interventions in order to optimize a patient's care path (using the number of future hospitalizations as an optimization criteria). Using fuzzy clustering, the method first creates patient clusters that are scored by the average number of future hospitalizations across the patients in the cluster. The method then allows for inspection of each patient's diagnostic codes and proposes a set of diagnostic changes (interventions) that will change a patient's cluster assignment to some cluster with less average hospitalizations. Finally, Dabek and Caban's work[151] offered a unique perspective for modeling patients' trajectories by modeling clinical path as automata and using grammar induction algorithm to reduce the graph to identify a common trajectory. Deterministic and non-deterministic finite-state automata were used (i.e. DFA and NFA) to represent patients' trajectories. Using data from patients who sustained mild traumatic brain injury (mTBI) and developed post-traumatic stress disorder (PTSD), the authors were able to identify the most common path patients took from brain injury to PTSD.

## 4. Conclusion

In this review, we presented a comprehensive summary of recent computational work focusing on the use of patient similarity assessment on longitudinal health data for a wide range of medical applications. These technologies allow for the realization of an early vision of decision support systems that attempted to query and compare historical medical information stored in electronic health record systems[152–154]. The underlying idea rests on the belief that identifying patient journeys that are similar to that of an index patient can be used to predict the clinical outcomes for the index patient and to suggest beneficial interventions. Moreover, patient similarity analysis has the potential to identify unknown disease mechanisms and outcomes. However, successfully mining the medical data remains a challenge, and as of now there is no consensus on the best methodology for patient similarity analysis. We recommend that development of robust computational frameworks to handle data abstraction, similarity metric learning, patient journey clustering and outcome prediction is essential to move towards clinical applications.

## 5. References

1. Minor, L. B. *Stanford Medicine Health Trends Report: Harnessing the Power of Data in Health*. https://med.stanford.edu/content/dam/sm/sm-news/documents/StanfordMedicineHealthTrendsWhitePaper2017.pdf (2017).
2. Frankovich, J., Longhurst, C. A. & Sutherland, S. M. Evidence-Based Medicine in the EMR Era. *N. Engl. J. Med.* **365**, 1758–1759 (2011).
3. Gombar, S., Callahan, A., Califf, R., Harrington, R. & Shah, N. H. It is time to learn from patients like mine. *npj Digit. Med.* **2**, 16 (2019).
4. Budrionis, A. & Bellika, J. G. The Learning Healthcare System: Where are we now? A systematic review. *Journal of Biomedical Informatics* vol. 64 87–92 (2016).
5. Esteva, A. *et al.* A guide to deep learning in healthcare. *Nat. Med.* **25**, 24–29 (2019).
6. Miotto, R., Wang, F., Wang, S., Jiang, X. & Dudley, J. T. Deep learning for healthcare: review, opportunities and challenges. *Brief. Bioinform.* **19**, 1236–1246 (2018).
7. Xiao, C., Choi, E. & Sun, J. Opportunities and challenges in developing deep learning models using electronic health records data: a systematic review. *J. Am. Med. Informatics Assoc.* **25**, 1419–1428 (2018).
8. Rajkomar, A., Dean, J. & Kohane, I. Machine Learning in Medicine. *N. Engl. J. Med.* **380**, 1347–1358 (2019).
9. Eraslan, G., Avsec, Ž., Gagneur, J. & Theis, F. J. Deep learning: new computational modelling techniques for genomics. *Nat. Rev. Genet.* 1 (2019) doi:10.1038/s41576-019-0122-6.
10. Topol, E. J. High-performance medicine: the convergence of human and artificial intelligence. *Nat. Med.* **25**, 44–56 (2019).
11. Sharafoddini, A., Dubin, J. A. & Lee, J. Patient Similarity in Prediction Models Based on Health Data: A Scoping Review. *JMIR Med. Informatics* **5**, e7 (2017).
12. Pinaire, J., Azé, J., Bringay, S. & Landais, P. Patient healthcare trajectory. An essential monitoring tool: a systematic review. *Heal. Inf. Sci. Syst.* **5**, 1–18 (2017).
13. Parimbelli, E., Marini, S., Sacchi, L. & Bellazzi, R. Patient similarity for precision medicine: A systematic review. *Journal of Biomedical Informatics* vol. 83 87–96 (2018).
14. Alaa, A. M., Yoon, J., Hu, S. & van der Schaar, M. Personalized Risk Scoring for Critical Care Patients using Mixtures of Gaussian Process Experts. (2016).
15. Suo, Q. *et al.* Personalized disease prediction using a CNN-based similarity learning method. in *2017 IEEE International Conference on Bioinformatics and Biomedicine (BIBM)* vol. nan 811–816 (IEEE, 2017).
16. Schulam, P., Wigley, F. & Saria, S. Clustering longitudinal clinical marker trajectories from electronic health data: Applications to phenotyping and endotype discovery. 2956–2964 (2015).
17. Tamang, S. & Parsons, S. Using semi-parametric clustering applied to electronic health record time series data. in *Proceedings of the 2011 workshop on Data mining for medicine and healthcare - DMMH '11* vol. nan 72 (ACM, 2011).
18. Jensen, A. B. *et al.* Temporal disease trajectories condensed from population-wide registry data covering 6.2 million patients. *Nat. Commun.* **5**, 4022 (2014).
19. Zhang, J., Kowsari, K., Harrison, J. H., Lobo, J. M. & Barnes, L. E. Patient2Vec: A Personalized Interpretable Deep Representation of the Longitudinal Electronic Health Record. *IEEE Access* (2018) doi:10.1109/ACCESS.2018.2875677.

20. Sun, J., Sow, D., Hu, J. & Ebadollahi, S. Localized Supervised Metric Learning on Temporal Physiological Data. in *2010 20th International Conference on Pattern Recognition* vol. nan 4149–4152 (IEEE, 2010).
21. Sun, J., Sow, D., Hu, J. & Ebadollahi, S. A System for Mining Temporal Physiological Data Streams for Advanced Prognostic Decision Support. in *2010 IEEE International Conference on Data Mining* vol. nan 1061–1066 (IEEE, 2010).
22. Rouzbahman, M. & Chignell, M. Predicting ICU Death with Summarized Data: The Emerging Health Data Search Engine. (2014).
23. Chung, I., Kim, S., Lee, J., Hwang, S. J. & Yang, E. Mixed Effect Composite RNN-GP: A Personalized and Reliable Prediction Model for Healthcare. (2018).
24. Dhamala, J., Azuh, E., Al-Dujaili, A., Rubin, J. & O'Reilly, U.-M. Multivariate Time-series Similarity Assessment via Unsupervised Representation Learning and Stratified Locality Sensitive Hashing: Application to Early Acute Hypotensive Episode Detection. (2018).
25. Lehman, L. H. *et al.* A Physiological Time Series Dynamics-Based Approach to Patient Monitoring and Outcome Prediction. *IEEE J. Biomed. Heal. Informatics* **19**, 1068–1076 (2015).
26. Beaulieu-Jones, B. K., Orzechowski, P. & Moore, J. H. Mapping Patient Trajectories using Longitudinal Extraction and Deep Learning in the MIMIC-III Critical Care Database. *Pac. Symp. Biocomput.* **23**, 123–132 (2018).
27. Kim, Y. B., Hemberg, E. & O'Reilly, U.-M. Collision frequency locality-sensitive hashing for prediction of critical events. in *2017 39th Annual International Conference of the IEEE Engineering in Medicine and Biology Society (EMBC)* vol. 2017 3088–3093 (IEEE, 2017).
28. Bonomi, L. & Jiang, X. Pattern Similarity in Time Interval Sequences. in *2018 IEEE International Conference on Healthcare Informatics (ICHI)* vol. nan 434–435 (IEEE, 2018).
29. Hu, J., Wang, F., Sun, J., Sorrentino, R. & Ebadollahi, S. A healthcare utilization analysis framework for hot spotting and contextual anomaly detection. *AMIA ... Annu. Symp. proceedings. AMIA Symp.* **2012**, 360–9 (2012).
30. Schulam, P. & Arora, R. Disease Trajectory Maps. (2016).
31. Johnson, A. E. W. *et al.* MIMIC-III, a freely accessible critical care database. *Sci. Data* **3**, (2016).
32. Petersen, R. C. *et al.* Alzheimer's Disease Neuroimaging Initiative (ADNI): Clinical characterization. *Neurology* **74**, 201–209 (2010).
33. Pollard, T. J. *et al.* The eICU collaborative research database, a freely available multi-center database for critical care research. *Sci. Data* **5**, (2018).
34. Marek, K. *et al.* The Parkinson Progression Marker Initiative (PPMI). *Progress in Neurobiology* vol. 95 629–635 (2011).
35. Silva, I., Moody, G., Scott, D. J., Celi, L. A. & Mark, R. G. Predicting in-hospital mortality of ICU patients: The PhysioNet/Computing in cardiology challenge 2012. in *Computing in Cardiology* vol. 39 245–248 (2012).
36. Voorhees, E. M. & Hersh, W. Overview of the TREC 2012 Medical Records Track. *Twenty-First Text Retr. Conf. (TREC 2012) Proc.* (2012).
37. Berka, P. ECML/PKDD 2002 discovery challenge, download data about hepatitis. in (2002).
38. Dua, D. & Graff, C. UCI Machine Learning Repository. *University of California, Irvine,*

*School of Information and Computer Sciences* http://archive.ics.uci.edu/ml (2017).
39. Kasabov, N. & Hu, Y. Integrated optimisation method for personalised modelling and case studies for medical decision support. *Int. J. Funct. Inform. Personal. Med.* **3**, 236–256 (2010).
40. Wivel, A. E., Lapane, K., Kleoudis, C., Singer, B. H. & Horwitz, R. I. Medicine Based Evidence for Individualized Decision Making: Case Study of Systemic Lupus Erythematosus. *Am. J. Med.* **130**, 1290-1297.e6 (2017).
41. Shknevsky, A., Shahar, Y. & Moskovitch, R. Consistent discovery of frequent interval-based temporal patterns in chronic patients' data. *J. Biomed. Inform.* **75**, 83–95 (2017).
42. Moskovitch, R. & Shahar, Y. Medical temporal-knowledge discovery via temporal abstraction. *AMIA ... Annu. Symp. proceedings. AMIA Symp.* **2009**, 452–6 (2009).
43. Moskovitch, R., Walsh, C., Wang, F., Hripcsak, G. & Tatonetti, N. Outcomes Prediction via Time Intervals Related Patterns. in *2015 IEEE International Conference on Data Mining* vol. nan 919–924 (IEEE, 2015).
44. Moskovitch, R., Choi, H., Hripcsak, G. & Tatonetti, N. Prognosis of Clinical Outcomes with Temporal Patterns and Experiences with One Class Feature Selection. *IEEE/ACM Trans. Comput. Biol. Bioinforma.* **14**, 555–563 (2017).
45. Moskovitch, R., Polubriaginof, F., Weiss, A., Ryan, P. & Tatonetti, N. Procedure prediction from symbolic Electronic Health Records via time intervals analytics. *J. Biomed. Inform.* **75**, 70–82 (2017).
46. Lin, J. & Li, Y. Finding structurally different medical data. in *2009 22nd IEEE International Symposium on Computer-Based Medical Systems* vol. nan 1–8 (IEEE, 2009).
47. Keogh, E., Lonardi, S. & Ratanamahatana, C. A. Towards parameter-free data mining. in *KDD-2004 - Proceedings of the Tenth ACM SIGKDD International Conference on Knowledge Discovery and Data Mining* 206–215 (2004). doi:10.1145/1014052.1014077.
48. Zhao, J., Papapetrou, P., Asker, L. & Boström, H. Learning from heterogeneous temporal data in electronic health records. *J. Biomed. Inform.* **65**, 105–119 (2017).
49. Ghosh, S., Feng, M., Nguyen, H. & Li, J. Hypotension Risk Prediction via Sequential Contrast Patterns of ICU Blood Pressure. *IEEE J. Biomed. Heal. Informatics* **20**, 1416–1426 (2016).
50. Morid, M. A., Sheng, O. R. L. & Abdelrahman, S. Leveraging Patient Similarity and Time Series Data in Healthcare Predictive Models. (2017).
51. Katsuki, T. *et al.* Risk Prediction of Diabetic Nephropathy via Interpretable Feature Extraction from EHR Using Convolutional Autoencoder. *Stud. Health Technol. Inform.* **247**, 106–110 (2018).
52. Zhang, Y., Zhou, H., Li, J., Sun, W. & Chen, Y. A Time-Sensitive Hybrid Learning Model for Patient Subgrouping. in *2018 International Joint Conference on Neural Networks (IJCNN)* vol. nan 1–8 (IEEE, 2018).
53. Mikalsen, K. Ø., Bianchi, F. M., Soguero-Ruiz, C. & Jenssen, R. Time Series Cluster Kernel for Learning Similarities between Multivariate Time Series with Missing Data. (2017).
54. Bianchi, F. M., Livi, L., Mikalsen, K. Ø., Kampffmeyer, M. & Jenssen, R. Learning representations for multivariate time series with missing data using Temporal Kernelized Autoencoders. (2018).

55. Hinton, G. E. & Salakhutdinov, R. R. Reducing the dimensionality of data with neural networks. *Science (80-. ).* **313**, 504–507 (2006).
56. Beaulieu-Jones, B. K. & Greene, C. S. Semi-supervised learning of the electronic health record for phenotype stratification. *J. Biomed. Inform.* **64**, 168–178 (2016).
57. Van Der Maaten, L. & Hinton, G. Visualizing data using t-SNE. *J. Mach. Learn. Res.* (2008).
58. Baytas, I. M. *et al.* Patient Subtyping via Time-Aware LSTM Networks. in *Proceedings of the 23rd ACM SIGKDD International Conference on Knowledge Discovery and Data Mining - KDD '17* 65–74 (ACM Press, 2017). doi:10.1145/3097983.3097997.
59. Henriques, J., Rocha, T., Paredes, S. & de Carvalho, P. Telehealth streams reduction based on pattern recognition techniques for events detection and efficient storage in EHR. in *2013 35th Annual International Conference of the IEEE Engineering in Medicine and Biology Society (EMBC)* vol. nan 7488–7491 (IEEE, 2013).
60. Montani, S., Portinale, L., Leonardi, G., Bellazzi, R. & Bellazzi, R. Case-based retrieval to support the treatment of end stage renal failure patients. *Artif. Intell. Med.* **37**, 31–42 (2006).
61. Sacchi, L., D'Ancona, G., Bertuzzi, F. & Bellazzi, R. Temporal clustering for blood glucose analysis in the ICU: identification of groups of patients with different risk profile. *Stud. Health Technol. Inform.* **160**, 1150–4 (2010).
62. Batal, I. *et al.* An efficient pattern mining approach for event detection in multivariate temporal data. *Knowl. Inf. Syst.* **46**, 115–150 (2016).
63. Saeed, M. & Mark, R. A novel method for the efficient retrieval of similar multiparameter physiologic time series using wavelet-based symbolic representations. *AMIA ... Annu. Symp. proceedings. AMIA Symp.* **2006**, 679–83 (2006).
64. Lara, J. A., Lizcano, D., Pérez, A. & Valente, J. P. A general framework for time series data mining based on event analysis: Application to the medical domains of electroencephalography and stabilometry. *J. Biomed. Inform.* **51**, 219–241 (2014).
65. Sheetrit, E. *et al.* Temporal Pattern Discovery for Accurate Sepsis Diagnosis in ICU Patients. (2017).
66. Baxter, R., Williams, G. & He, H. Feature Selection for Temporal Health Records. *Adv. Knowl. Discov. Data Min.* **2035**, 198–209 (2001).
67. Berlingerio, M., Bonchi, F., Giannotti, F. & Turini, F. Mining Clinical Data with a Temporal Dimension: A Case Study. in *2007 IEEE International Conference on Bioinformatics and Biomedicine (BIBM 2007)* vol. nan 429–436 (IEEE, 2007).
68. Bock, C. *et al.* Association mapping in biomedical time series via statistically significant shapelet mining. *Bioinformatics* **34**, i438–i446 (2018).
69. Guéguin, M. *et al.* Exploring time series retrieved from cardiac implantable devices for optimizing patient follow-up. *IEEE Trans. Biomed. Eng.* **55**, 2343–52 (2008).
70. Wang, X., Wang, Y., Gao, C., Lin, K. & Li, Y. Automatic Diagnosis With Efficient Medical Case Searching Based on Evolving Graphs. *IEEE Access* **6**, 53307–53318 (2018).
71. Zhang, S., Liu, L., Li, H., Xiao, Z. & Cui, L. MTPGraph: A data-driven approach to predict medical risk based on temporal profile graph. *Proc. - 15th IEEE Int. Conf. Trust. Secur. Priv. Comput. Commun. 10th IEEE Int. Conf. Big Data Sci. Eng. 14th IEEE Int. Symp. Parallel Distrib. Proce* **nan**, 1174–1181 (2016).
72. Liu, C., Wang, F., Hu, J. & Xiong, H. Temporal Phenotyping from Longitudinal Electronic Health Records: A Graph Based Framework. in *Proceedings of the 21th ACM SIGKDD International Conference on Knowledge Discovery and Data Mining -*

*KDD '15* vol. nan 705–714 (ACM Press, 2015).
73. Chan, L., Chan, T., Cheng, L. & Mak, W. Machine learning of patient similarity: A case study on predicting survival in cancer patient after locoregional chemotherapy. in *2010 IEEE International Conference on Bioinformatics and Biomedicine Workshops (BIBMW)* 467–470 (IEEE, 2010). doi:10.1109/BIBMW.2010.5703846.
74. Gottlieb, A., Stein, G. Y., Ruppin, E., Altman, R. B. & Sharan, R. A method for inferring medical diagnoses from patient similarities. *BMC Med.* **11**, 194 (2013).
75. Wang, Y., Tian, Y., Tian, L.-L., Qian, Y.-M. & Li, J.-S. An Electronic Medical Record System with Treatment Recommendations Based on Patient Similarity. *J. Med. Syst.* **39**, 55 (2015).
76. Sun, L., Liu, C., Guo, C., Xiong, H. & Xie, Y. Data-driven Automatic Treatment Regimen Development and Recommendation. *Proc. 22nd ACM SIGKDD Int. Conf. Knowl. Discov. Data Min. - KDD '16* **nan**, 1865–1874 (2016).
77. Zhang, J., Huang, J. X., Guo, J. & Xu, W. Promoting electronic health record search through a time-aware approach. in *2013 IEEE International Conference on Bioinformatics and Biomedicine* vol. nan 593–596 (IEEE, 2013).
78. Lee, J., Maslove, D. M. & Dubin, J. A. Personalized Mortality Prediction Driven by Electronic Medical Data and a Patient Similarity Metric. *PLoS One* **10**, e0127428 (2015).
79. Garcelon, N. *et al.* Finding patients using similarity measures in a rare diseases-oriented clinical data warehouse: Dr. Warehouse and the needle in the needle stack. *J. Biomed. Inform.* **73**, 51–61 (2017).
80. Kim, Y. B., O'Reilly, U.-M., O'Reilly, Y. B. K. U., Kim, Y. B. & O'Reilly, U.-M. Large-scale physiological waveform retrieval via locality-sensitive hashing. in *2015 37th Annual International Conference of the IEEE Engineering in Medicine and Biology Society (EMBC)* vol. nan 5829–5833 (IEEE, 2015).
81. Hielscher, T., Völzke, H., Papapetrou, P. & Spiliopoulou, M. Discovering, selecting and exploiting feature sequence records of study participants for the classification of epidemiological data on hepatic steatosis. in *Proceedings of the 33rd Annual ACM Symposium on Applied Computing - SAC '18* vol. nan 6–13 (ACM Press, 2018).
82. Hielscher, T., Spiliopoulou, M., Volzke, H. & Kuhn, J.-P. Using Participant Similarity for the Classification of Epidemiological Data on Hepatic Steatosis. in *2014 IEEE 27th International Symposium on Computer-Based Medical Systems* vol. nan 1–7 (IEEE, 2014).
83. Sha, Y., Venugopalan, J. & Wang, M. D. A Novel Temporal Similarity Measure for Patients Based on Irregularly Measured Data in Electronic Health Records. in *Proceedings of the 7th ACM International Conference on Bioinformatics, Computational Biology, and Health Informatics - BCB '16* vol. nan 337–344 (ACM Press, 2016).
84. Syed, H. & Das, A. K. Temporal Needleman-Wunsch. in *2015 IEEE International Conference on Data Science and Advanced Analytics (DSAA)* 1–9 (IEEE, 2015). doi:10.1109/DSAA.2015.7344785.
85. Hajihashemi, Z. & Popescu, M. Detection of abnormal sensor patterns in eldercare. in *2013 E-Health and Bioengineering Conference (EHB)* vol. nan 1–4 (IEEE, 2013).
86. Hajihashemi, Z. & Popescu, M. A Multidimensional Time-Series Similarity Measure With Applications to Eldercare Monitoring. *IEEE J. Biomed. Heal. Informatics* **20**, 953–962 (2016).

87. Zhang, J., Xu, W., Guo, J. & Gao, S. A temporal model in Electronic Health Record search. *Knowledge-Based Syst.* **126**, 56–67 (2017).
88. Zhang, J., Xu, W. & Guo, J. A time-aware approach for boosting medical records search. in *2016 Digital Media Industry & Academic Forum (DMIAF)* vol. nan 99–102 (IEEE, 2016).
89. Aronson, A. R. & Lang, F. M. An overview of MetaMap: Historical perspective and recent advances. *J. Am. Med. Informatics Assoc.* **17**, 229–236 (2010).
90. Tsevas, S. & Iakovidis, D. K. Dynamic time warping fusion for the retrieval of similar patient cases represented by multimodal time-series medical data. in *Proceedings of the 10th IEEE International Conference on Information Technology and Applications in Biomedicine* vol. nan 1–4 (IEEE, 2010).
91. Salgado, C. M., Ferreira, M. C. & Vieira, S. M. Mixed Fuzzy Clustering for Misaligned Time Series. *IEEE Trans. Fuzzy Syst.* **25**, 1777–1794 (2017).
92. Goyal, D., Syed, Z. & Wiens, J. Clinically Meaningful Comparisons Over Time: An Approach to Measuring Patient Similarity based on Subsequence Alignment. (2018).
93. Hoogendoorn, M. *et al.* Prediction using patient comparison vs. modeling: A case study for mortality prediction. in *2016 38th Annual International Conference of the IEEE Engineering in Medicine and Biology Society (EMBC)* vol. nan 2464–2467 (IEEE, 2016).
94. Sun, J., Wang, F., Hu, J. & Edabollahi, S. Supervised patient similarity measure of heterogeneous patient records. *ACM SIGKDD Explor. Newsl.* **14**, 16 (2012).
95. Wang, F., Sun, J. & Ebadollahi, S. Integrating distance metrics learned from multiple experts and its application in patient similarity assessment. *Proc. 11th SIAM Int. Conf. Data Mining, SDM 2011* 59–70 (2011) doi:10.1137/1.9781611972818.6.
96. Ebadollahi, S. *et al.* Predicting Patient's Trajectory of Physiological Data using Temporal Trends in Similar Patients: A System for Near-Term Prognostics. *AMIA ... Annu. Symp. proceedings. AMIA Symp.* **2010**, 192–6 (2010).
97. Zhan, M., Cao, S., Qian, B., Chang, S. & Wei, J. Low-rank sparse feature selection for patient similarity learning. *Proc. - IEEE Int. Conf. Data Mining, ICDM* 1335–1340 (2017) doi:10.1109/ICDM.2016.93.
98. Wang, F., Hu, J. & Sun, J. Medical prognosis based on patient similarity and expert feedback. in *Proceedings - International Conference on Pattern Recognition* (2012).
99. Ng, K., Sun, J., Hu, J. & Wang, F. Personalized Predictive Modeling and Risk Factor Identification using Patient Similarity. *AMIA Jt. Summits Transl. Sci. proceedings. AMIA Jt. Summits Transl. Sci.* **2015**, 132–6 (2015).
100. Zhu, Z. *et al.* Measuring Patient Similarities via a Deep Architecture with Medical Concept Embedding. in *2016 IEEE 16th International Conference on Data Mining (ICDM)* vol. nan 749–758 (IEEE, 2016).
101. Suo, Q. *et al.* Deep Patient Similarity Learning for Personalized Healthcare. *IEEE Trans. Nanobioscience* **17**, 219–227 (2018).
102. Che, C. *et al.* An RNN architecture with dynamic temporal matching for personalized predictions of Parkinson's disease. *Proc. 17th SIAM Int. Conf. Data Mining, SDM 2017* 198–206 (2017) doi:10.1137/1.9781611974973.23.
103. Robert, P. & Escoufier, Y. A Unifying Tool for Linear Multivariate Statistical Methods: The RV- Coefficient. *Appl. Stat.* **25**, 257 (1976).
104. Josse, J. & Holmes, S. Measuring multivariate association and beyond. *Stat. Surv.* **10**, 132–167 (2016).

105. Hirano, S., Xiaoguang Sun & Tsumoto, S. Analysis of time-series medical databases using multiscale structure matching and rough sets-based clustering technique. in *10th IEEE International Conference on Fuzzy Systems. (Cat. No.01CH37297)* vol. 2 1547–1550 (IEEE, 2001).
106. Hirano, S. & Tsumoto, S. Mining similar temporal patterns in long time-series data and its application to medicine. in *2002 IEEE International Conference on Data Mining, 2002. Proceedings.* vol. nan 219–226 (IEEE Comput. Soc, 2002).
107. Hirano, S. & Tsumoto, S. Multiscale Comparison of Temporal Patterns in Time-Series Medical Databases. *KDD, LNAI 2431* **nan**, 188–199 (2002).
108. Hirano, S. & Tsumoto, S. Multiscale analysis of long time-series medical databases. *AMIA ... Annu. Symp. proceedings. AMIA Symp.* 289–93 (2003).
109. Tsumoto, S. & Hirano, S. Automated discovery of chronological patterns in long time-series medical datasets. *Int. J. Intell. Syst.* **20**, 737–757 (2005).
110. Hirano, S. & Tsumoto, S. Strucutral Comparison and Cluster Analysis of Time-Series Medical Data. in *2005 IEEE International Conference on Systems, Man and Cybernetics* vol. 2 1506–1511 (IEEE, 2005).
111. Hirano, S. & Tsumoto, S. Clustering Time-Series Medical Databases Based on the Improved Multiscale Matching. in *Lecture Notes in Computer Science book series (LNCS, volume 3488)* vol. nan 612–621 (Springer-Verlag, 2005).
112. Hirano, S. & Tsumoto, S. Cluster Analysis of Time-Series Medical Data Based on the Trajectory Representation and Multiscale Comparison Techniques. in *Sixth International Conference on Data Mining (ICDM'06)* vol. nan 896–901 (IEEE, 2006).
113. Hirano, S. & Tsumoto, S. Grouping Similar Trajectories in Hospital Labo ratory Data. in *2007 IEEE/ICME International Conference on Complex Medical Engineering* vol. nan 1933–1939 (IEEE, 2007).
114. Hira, S. & Tsumoto, S. Identifying Exacerbating Cases in Chronic Diseases Based on the Cluster Analysis of Trajectory Data on Laboratory Examinations. in *Seventh IEEE International Conference on Data Mining Workshops (ICDMW 2007)* vol. nan 151–156 (IEEE, 2007).
115. Tsumoto, S. & Hirano, S. Mining Trajectories of Laboratory Data using Multiscale Matching and Clustering. in *2008 21st IEEE International Symposium on Computer-Based Medical Systems* vol. nan 626–631 (IEEE, 2008).
116. Hirano, S. & Tsumoto, S. Curvature Maxima-based Trajectories Mining. in *2010 IEEE International Conference on Data Mining Workshops* vol. nan 257–264 (IEEE, 2010).
117. Chao, G., Mao, C., Wang, F., Zhao, Y. & Luo, Y. Supervised Nonnegative Matrix Factorization to Predict ICU Mortality Risk. (2018) doi:arXiv:1809.10680v2.
118. Wang, F., Lee, N., Hu, J., Sun, J. & Ebadollahi, S. Towards heterogeneous temporal clinical event pattern discovery. in *Proceedings of the 18th ACM SIGKDD international conference on Knowledge discovery and data mining - KDD '12* 453 (ACM Press, 2012). doi:10.1145/2339530.2339605.
119. Wang, F. *et al.* A Framework for Mining Signatures from Event Sequences and Its Applications in Healthcare Data. *IEEE Trans. Pattern Anal. Mach. Intell.* **35**, 272–285 (2013).
120. Titus, A., Faill, R. & Das, A. Automatic Identification of Co-Occuring Patient Events. in *Proceedings of the 7th ACM International Conference on Bioinformatics, Computational Biology, and Health Informatics - BCB '16* vol. nan 579–586 (ACM Press, 2016).

121. Liu, J., Brodley, C. E., Healy, B. C. & Chitnis, T. Removing confounding factors via constraint-based clustering: An application to finding homogeneous groups of multiple sclerosis patients. *Artif. Intell. Med.* **65**, 79–88 (2015).
122. Sarkar, M. Measures of ruggedness using fuzzy-rough sets and fractals: applications in medical time series. in *2001 IEEE International Conference on Systems, Man and Cybernetics. e-Systems and e-Man for Cybernetics in Cyberspace (Cat.No.01CH37236)* vol. 3 1514–1519 (IEEE, 2002).
123. Pimentel, M. A. F., Clifton, D. A., Clifton, L., Watkinson, P. J. & Tarassenko, L. Modelling physiological deterioration in post-operative patient vital-sign data. *Med. Biol. Eng. Comput.* **51**, 869–77 (2013).
124. Li, X., Liu, H., Mei, J., Yu, Y. & Xie, G. Mining Temporal and Data Constraints Associated with Outcomes for Care Pathways. *Stud. Health Technol. Inform.* **216**, 711–715 (2015).
125. Combi, C., Mantovani, M. & Sala, P. Discovering Quantitative Temporal Functional Dependencies on Clinical Data. in *2017 IEEE International Conference on Healthcare Informatics (ICHI)* vol. nan 248–257 (IEEE, 2017).
126. Afshar, A. *et al.* COPA: Constrained PARAFAC2 for Sparse & Large Datasets. (2018) doi:arXiv:1803.04572v2.
127. Haimowitz, I. J., Le, P. P. & Kohane, I. S. Clinical monitoring using regression-based trend templates. *Artif. Intell. Med.* **7**, 473–496 (1995).
128. Galagali, N. & Xu-Wilson, M. Patient Subtyping with Disease Progression and Irregular Observation Trajectories. (2018).
129. Huang, Z., Dong, W., Wang, F. & Duan, H. Medical Inpatient Journey Modeling and Clustering: A Bayesian Hidden Markov Model Based Approach. *AMIA ... Annu. Symp. proceedings. AMIA Symp.* **2015**, 649–58 (2015).
130. Goodwin, T. & Harabiu, S. M. A Predictive Chronological Model of Multiple Clinical Observations. in *2015 International Conference on Healthcare Informatics* vol. nan 253–262 (IEEE, 2015).
131. Li, X. *et al.* Automatic variance analysis of multistage care pathways. *Stud. Health Technol. Inform.* **205**, 715–719 (2014).
132. Zhang, Y. & Padman, R. Innovations in chronic care delivery using data-driven clinical pathways. *Am. J. Manag. Care* **21**, e661-8 (2015).
133. Lehman, L. H. *et al.* Tracking progression of patient state of health in critical care using inferred shared dynamics in physiological time series. in *2013 35th Annual International Conference of the IEEE Engineering in Medicine and Biology Society (EMBC)* vol. nan 7072–7075 (IEEE, 2013).
134. Lehman, L. H., Nemati, S., Moody, G. B., Heldti, T. & Mark, R. G. Uncovering Clinical Significance of Vital Sign Dynamics in Critical Care. in *Computing in cardiology* vol. 41 1141–1144 (2014).
135. James, G. Principal component models for sparse functional data. *Biometrika* **87**, 587–602 (2000).
136. Hoffman, M. D. *et al. Stochastic Variational Inference*. *Journal of Machine Learning Research* vol. 14 (2013).
137. Zeldow, B., Flory, J., Stephens-Shields, A., Raebel, M. & Roy, J. Outcome identification in electronic health records using predictions from an enriched Dirichlet process mixture. (2018).
138. Alaa, A. M., Yoon, J., Hu, S. & van der Schaar, M. Personalized Risk Scoring for Critical

Care Prognosis Using Mixtures of Gaussian Processes. *IEEE Trans. Biomed. Eng.* **65**, 207–218 (2018).
139. Alaa, A. M., Yoon, J., Hu, S. & van der Schaar, M. Personalized Risk Scoring for Critical Care Patients using Mixtures of Gaussian Process Experts. **48**, (2016).
140. Alaa, A. M., Yoon, J., Hu, S. & van der Schaar, M. Individualized Risk Prognosis for Critical Care Patients: A Multi-task Gaussian Process Model. (2017).
141. Marlin, B. M., Kale, D. C., Khemani, R. G. & Wetzel, R. C. Unsupervised pattern discovery in electronic health care data using probabilistic clustering models. in *Proceedings of the 2nd ACM SIGHIT symposium on International health informatics - IHI '12* 389 (ACM Press, 2012). doi:10.1145/2110363.2110408.
142. Suresh, H., Gong, J. J. & Guttag, J. Learning Tasks for Multitask Learning: Heterogenous Patient Populations in the ICU. (2018) doi:10.1145/3219819.3219930.
143. Tran, T., Nguyen, T. D., Phung, D. & Venkatesh, S. Learning vector representation of medical objects via EMR-driven nonnegative restricted Boltzmann machines (eNRBM). *J. Biomed. Inform.* **54**, 96–105 (2015).
144. Fisher, C. K., Smith, A. M., Walsh, J. R. & Diseases, the C. A. M. Deep learning for comprehensive forecasting of Alzheimer's Disease progression. (2018).
145. Rao, B. R., Sandilya, S., Niculescu, R., Germond, C. & Goel, A. Mining time-dependent patient outcomes from hospital patient records. *Proceedings. AMIA Symp.* 632–6 (2002).
146. Zalewski, A., Long, W., Johnson, A. E. W., Mark, R. G. & Lehman, L. W. H. Estimating patient's health state using latent structure inferred from clinical time series and text. in *2017 IEEE EMBS International Conference on Biomedical and Health Informatics, BHI 2017* vol. nan 449–452 (IEEE, 2017).
147. Westergaard, D., Moseley, P., Sørup, F. K. H., Baldi, P. & Brunak, S. Population-wide analysis of differences in disease progression patterns in men and women. *Nat. Commun.* (2019) doi:10.1038/s41467-019-08475-9.
148. Lademann, M., Lademann, M., Boeck Jensen, A. & Brunak, S. Incorporating symptom data in longitudinal disease trajectories for more detailed patient stratification. *Int. J. Med. Inform.* (2019) doi:10.1016/j.ijmedinf.2019.06.003.
149. Oh, W. *et al.* Type 2 Diabetes Mellitus Trajectories and Associated Risks. *Big data* **4**, 25–30 (2016).
150. Al-Mardini, M. *et al.* Reduction of Hospital Readmissions through Clustering Based Actionable Knowledge Mining. in *2016 IEEE/WIC/ACM International Conference on Web Intelligence (WI)* vol. nan 444–448 (IEEE, 2016).
151. Dabek, F. & Caban, J. J. A grammar-based approach to model the patient's clinical trajectory after a mild traumatic brain injury. in *2015 IEEE International Conference on Bioinformatics and Biomedicine (BIBM)* vol. nan 723–730 (IEEE, 2015).
152. Califf, R. M. & Rosati, R. A. The doctor and the computer. *West. J. Med.* **135**, 321–3 (1981).
153. Schultz, J. R., Cantrill, S. V. & Morgan, K. G. An initial operational problem oriented medical record system. in *Proceedings of the May 18-20, 1971, spring joint computer conference on - AFIPS '71 (Spring)* vol. nan 239 (ACM Press, 1971).
154. CHUTE, C. G. Clinical Data Retrieval and Analysis. I've Seen a Case Like That Before. *Ann. N. Y. Acad. Sci.* **670**, 133–140 (1992).